\documentclass{article}
\usepackage{spconf,amsmath,amssymb,graphicx}
\usepackage{dsfont}
\usepackage{siunitx}
\usepackage{float}
\usepackage{hyperref}

\title{The Zero Resource Speech Challenge 2017}
\name{E. Dunbar{$^1$}, X-N. Cao{$^1$}, J. Benjumea{$^1$}, J. Karadayi{$^1$}, M. Bernard{$^1$}, L. Besacier{$^2$}, X. Anguera{$^3$}, E. Dupoux{$^1$}}
\address{{$^1$}ENS, EHESS, PSL Research University, CNRS, INRIA\\
	{$^2$} LIG, Univ. Grenoble Alpes, CNRS, Grenoble INP, France\\
	{$^3$} Elsa Corp., Lisbon, Portugal}

\begin{document}
%\ninept
%
\maketitle

%
%%%Abstract
\begin{abstract}
  We describe a new challenge aimed at discovering subword and word units from raw speech. 
This challenge is the followup to the Zero Resource Speech Challenge 2015. It aims at constructing systems that generalize across languages and adapt to new speakers. The design features and evaluation metrics of the challenge are presented and the results of seventeen models are discussed.
\end{abstract}

\section{Introduction}
\label{sec:intro}
Traditional speech technology relies on linguistic expertise and textual information to build acoustic and language models. Recent work has begun to mitigate the need for expert knowledge,
%phonetic dictionaries and expert knowledge, 
having succeeded at training supervised systems using textual transcripts only \cite{baidudeepspeech,miao2015eesen}. But for the majority of the world's languages, even textual transcripts are sparse or non-existent, and for many languages, the lack of standard orthography would reduce the utility of such transcripts greatly. This is why there is increasing interest in a radically different approach, the so-called \textit{zero resource setting} \cite{glass_2012,jansen_2013}. The aim is to construct speech systems without any textual or linguistic resource. The fact that young children become fluent communicators in their native language(s) before they learn to read and write, and with minimal or no expert supervision, demonstrates that such a system is possible.
%The Zero Resource Speech Challenge is an effort at exploratory work in ``resourceless'' learning of speech: learning which works from the speech signal without any additional linguistic expertise (orthography, linguistic transcriptions, language identification) that would not be systematically available to human children learning a first language. The scientific question this type of research addresses is to advance our understanding the problem of early language acquisition, with a modest (but still distant) goal: a system which, after learning autonomously in this way, yields an end-to-end spoken dialogue system with human-like performance---a functional model of a human learning a native language. There are also several practical benefits of this kind of research. Traditional speech and language technologies are trained with massive amounts of textual information. However, many of the world's languages do not have textual resources or even a reliable orthography. 

Systems constructed with zero expert resources have at least three kinds of applications. They can form the basis for speech services for the perhaps billions of speakers of languages without readily available linguistic or textual resources in sufficient quantities. They can also help field linguists document endangered languages, by providing tools to semi-automatically analyze and annotate audio recordings using automatically discovered linguistic units (phonemes, lexicon, grammar) \cite{addaBulbSLTU2016}. In this case, a realistic scenario is to process raw speech and its translation in a well-resourced language \cite{duong2016,berard_2016,besacier_2006}. Finally, to the extent that they mimic language development, these models can be used to evaluate scientific claims about development, and act as quantitative tools for psychologists and clinicians interested in the impact of sociolinguistic variations in input on subsequent normal or abnormal language and cognitive development \cite{dupoux2016}.

As the previous Zero Resource Speech Challenge \cite{versteegh_2015,versteegh_2016}, the 2017 installment consists of two different sub-challenges. Participants can submit systems that do either \emph{subword modelling} (track one) or \emph{spoken term discovery} (track two). Subword modelling means constructing a representation of speech sounds; we say ``subword'' instead of ``phoneme'' or ``sound'' because the criterion for success does not demand any particular temporal window size. The requirement is to have a representation of the speech signal that is robust to %within- and between-talker 
talker variation, and makes the distinctions necessary to distinguish words.
%and which makes the distinctions necessary to recognize words. 
Spoken term discovery means discovering recurring speech fragments in the audio: the system must output time-stamps delimiting stretches of speech, associated with class labels, ideally corresponding to real words in the language. The systems must be able to do their task without linguistic resources. The 2017 Zero Resource Speech Challenge\footnote{http://zerospeech.com/2017} has two main innovations compared to the 2015 iteration: (1) it tests how well systems' architectures and hyperparameters generalize to unseen languages (the systems are developed with three languages and tested on two `surprise' languages); (2) it tests how well the trained systems' parameters adapt to unseen talkers (varying the amount of speech available for each talker).

\section{The challenge}
\label{sec:overview}

\subsection{Track one: Subword modelling}

The goal of subword modelling is to learn representations appropriate to the sounds of the language (phones or phonemes). As adults, we easily perceive differences between individual phonemes in our native language, substantially better than in non-native languages \cite{lisker1964cross}. This so-called \textit{perceptual tuning} develops substantially during the first year of life \cite{werker1984cross,kuhl1991human}. The input to subword modelling is a set of acoustic features encoding continuous speech. The result is a transformation that applies to new samples of running speech, yielding a new representation (continuous or discrete) of that speech. After training in a particular language from unlabelled speech, the representation is computed for a test set in the same language on novel speakers, to be evaluated on phoneme discrimination based on gold-standard annotations.

\subsection{Track two: Spoken term discovery}

Intuitively, the goal of spoken term discovery is to find words in the speech stream---just as the infant learns the words of its language by listening. The input to candidate systems is a series of speech features. The output is a set of boundaries delimiting the start and end of proposed word tokens discovered in the speech, and category labels indicating proposed word types. These boundaries may, but need not, constitute an exhaustive parse of the speech. The evaluation we apply is a set of scores measuring different aspects of the alignment with the words in the gold-standard transcription. % (see below under \emph{Evaluation}). 
%Some systems may make use of a training phase (for example, XXX-CITATION-OKKO), while others may not (for example, XXX-CITATION-AREN). 
As is customary in the field of word segmentation, we do not provide a separate test set for this track; we rely on the surprise languages to assess possible hyperparameter overfitting.

%Although it is possible to run these two tracks independently, one could propose systems with interactions between the two tracks (e.g., using the representations of track 1 to improve track 2 or vice versa).
%Recent developmental psycholinguistic literature suggests that specialized native phoneme discrimination (indicating improved subword representations) begins to emerge in infants around the same time as the recognition of native language words (XXX). This runs contrary to previously widely held views that phonetic category learning is complete before word learning starts. Therefore, a system that uses untransformed acoustic features as the input to spoken term discovery---rather than a subword representation trained on the language---would not be inconsistent with the current understanding of human development.

\subsection{Generalization across languages}
This challenge is explicitly set up as a `learning to learn' or meta-learning problem: participants select the best architectures and tune their hyperparameters on three development languages on which the evaluation software is provided for a test set. The higher level test is performed on two new surprise languages for which the evaluation is withheld. Participants submit the output of their previously optimized systems for these two corpora, and the organizers perform the evaluation. 
%The challenge explicitly precludes the usage of labels to train the systems. To facilitate development, however, we nevertheless provide participants with access to the evaluation software (which uses knowledge of the gold transcriptions in order to evaluate the systems) for 3 languages which work as a development set. The systems, however, are compared on two surprise languages for which the evaluation and labels are withheld.
Each submission (maximum five per team) is automatically evaluated and made public. 

\subsection{Adaptation to new speakers}
The subword modeling task is explicitly set up as a talker adaptation problem. The training data is constructed from a skewed distribution of speakers, and the test set consists only of new speakers appearing in files of various durations (from 1 second to two minutes). 
%The evaluation of subword modelling is similar to the previous challenge: a phoneme discrimination score. But here, the test sets on which the systems are evaluated consist entirely of unseen talkers---not present in training---for both the development and the surprise corpora. These talkers are presented in files which can be very short (from one second to two minutes). 
This enables us to evaluate models' ability to generalize and adapt to new speakers on the basis of limited evidence. %See \emph{Data sets} and \emph{Evaluation} for more details.

\section{Data sets}
\label{sec:data}

Two groups of data sets are provided as part of the challenge:
the \emph{development data} and the \emph{surprise data}. The development data consists of corpora from three different languages (English, French and Mandarin). Each corpus comes with software that does the evaluation for the two tracks. Challenge participants are encouraged to use these resources to tune their hyperparameters using a cross validation approach to maximize generalizability. Each corpus is split into a train and a test set; the subword modelling evaluations are run on the test sets, and the spoken term discovery evaluations on the training sets. The participants then have to submit their systems and their output on all  the data  sets for independent evaluation (run automatically upon submission). The surprise data consists of corpora from two new languages, the identity of which was not revealed to participants. Participants receive only the speech corpora, with no additional resources. As before, each contains a training set (which participants can use to train their systems) and a test set for track one evaluation.

The amount of data in the training part of the development data sets varies from 2.5 to 45 hours (see Table \ref{tab:data}), to ensure that systems can work both with limited data and with large data sets. The statistics of the two surprise languages fall between these two extremes. 
%The division into development data and surprise data implements the idea of generalization across languages, which is intended to enforce the requirement of a truly zero resource system. The goal is to arrive at a system that yields good performance  without any hyperparameter tuning---more precisely, without tuning of any parameters in any way that demands linguistic resources. Furthermore, each of the corpora (both development and surprise) is set up so that generalization to new talkers is necessary: the test sets all consist of talkers that are novel, not present in the training set. Talker labels are provided for the training set only. 
The distribution of speakers in the training sets is shaped to reflect what is typically found in natural language acquisition settings: there is a ``family''---a small number of speakers that make up a large proportion of the total speech---and a set of ``outsiders''---a larger number of speakers that each appear in smaller proportions. The test sets consist of many short files, are organized into subsets of differing length (1s, 10s and 120s).
%This organization is illustrated in Figure \ref{fig:dataorg}. 

%      \begin{figure}
%        \centering
%        \includegraphics[width=0.6\linewidth]{zr2speakers.png}
%        \caption{Organization of data sets.}
%        \label{fig:dataorg}
%      \end{figure}

\begin{table*}[t]
\caption{\label{tab:data} Corpus statistics}
\centering

\resizebox{1.6\columnwidth}{!}{
\begin{tabular}{ l l l l l l l | l l }
	  & \multicolumn{6}{c}{Training} &  \multicolumn{2}{|c}{Test} \\
	   & \multicolumn{2}{c}{Relatives} &  \multicolumn{2}{c}{Outsiders}  &  \multicolumn{2}{c|}{Total}   \\ \hline
	   & \#speakers & duration/ speaker &  \#speakers & duration/ speaker &  duration & \#words &  number of files &  duration (min)  \\ \hline
	 \textbf{Development} \\ \hline
	 English & 9 & 165-220min &  60 & 10min & 45h & 370k  & 30658 & 1634 \\
	 French & 10 & 110-195min &  18 & 10min & 24h & 220k  & 23765 & 1061 \\
	 Mandarin & 4 & 20-25min &  8 & 10min &  2h30 & 20k  & 25383 & 1522 \\\hline
	 \textbf{Surprise} \\ \hline
	 S1 -- German & 10 & 85-150min &  20 & 10min &  25h & 213k  & 15243 & 687\\
	 S2 -- Wolof & 4 & 37-42min &  10 & 10min &  4h & 31k  & 7201 & 354 \\
\end{tabular}
}
\end{table*}

The English and French corpora were taken from the LibriVox collection of audio books\footnote{http://librivox.org/} and phone force-aligned using Kaldi \cite{Povey_ASRU2011_2011}. The Mandarin corpus is the one described in \cite{mandarinpaper}, force-aligned using Kaldi. The first surprise language is German, and was taken from LibriVox and force-aligned using Kaldi as well. The second surprise language is Wolof, and is the corpus described in \cite{wolofpaper}.

\begin{table*}[t!]
\caption{\label{tab:track1res} Track 1 within and between talker ABX scores for three test files durations, for the development (average) and surprise languages, for the 14 submitted systems (the mean performance on the surprise set is in square brackets). The three supervised systems are listed separately. See \url{http://zerospeech.com/2017} for the full table including development languages.}

\centering
\resizebox{2.05\columnwidth}{!}{

\begin{tabular}{ l  | l   l  l  l | l  l  l | l l l}
%     &  & \multicolumn{6}{c}{Surprise} \\
	 \multicolumn{2}{c}{} &\multicolumn{3}{c}{Avg. Development}  &  \multicolumn{3}{c}{S1 -- German} & \multicolumn{3}{c}{S2 -- Wolof}  \\
	 \multicolumn{2}{c}{} &1 s&10 s& 2 min&   1 s & 10 s & 2 min  & 1 s & 10 s & 2 min \\ \hline
	\textbf{Baseline} [18.84]
 & Within & 12 & 12.1 & 12.1 & 14.2 & 13.3 & 12.9 & 14.1 & 14.3 & 14.1 \\
	 & Between  & 23.3 & 23.4 & 23.3 & 27.5 & 27.3 & 27.1 & 30 & 29.5 & 29.5   \\ \hline
	\textbf{Topline} [7.19]
 & Within & 8 & 5.4 & 5.3 & 8.7 & 7.1 & 7.0 & 6.6 & 4.6 & 3.4   \\ 
	 & Between  & 9.6 & 7.2  & 6.9 & 12.8 & 10.5 & 10.4 & 7.1 & 3.6 & 4.3  \\\hline\hline
 
	\textbf{H} [9.16]%: \#1]
     & Within & 8.5 & 7.6 & 7.4 & 6.5 & 5.6 & 5.3 & 10.9 & 8.8 & 8.4\\ 
	 & Between  & 10.8 & 9.3 & 9.0 & 11.9 & 10.0 & 9.7 & 13.0 & 10.0 & 9.9 \\\hline	 
	\textbf{P1} [14.96]%: \#11]
     & Within & 10.9 & 8.8 & 8.7 & 8.9 & 6.7 & 6.4 & 13.3 & 11.9 & 11.8\\ 
	 & Between  & 17.5 & 15.8 & 15.7 & 19.4 & 16.2 & 15.9 & 22.8 & 23.1 & 23.1 \\
	\textbf{P2} [14.9]%4: \#10]
     & Within & 10.8 & 8.7 & 8.5 & 8.8 & 6.6 & 6.3 & 13.1 & 11.7 & 11.7\\ 
	 & Between  & 17.5 & 15.8 & 15.7 & 19.2 & 16.3 & 16.0 & 23.3 & 23.3 & 23.1\\
	 \hline 
	\textbf{C1} [11.95]%: \#4]
     & Within  & 10.0 & 8.4 & 8.3 & 7.6 & 6.22 & 6.3 & 11.7 & 9.9 & 9.8\\ 
	 & Between  & 14.5  & 12.8 & 12.5 & 15.5 & 12.9 & 12.7 & 17.6 & 16.9 & 16.3 \\
	\textbf{C2} [11.24]%: \#2]
     & Within & 10.1 & 8.5 & 8.4 & 7.6 & 6.2 & 6.1 & 11.6 & 9.8 & 9.6\\ 
	 & Between &  13.9 & 11.9 & 11.7 & 14.7 & 11.7 & 11.6 & 16.9 & 14.7 & 14.4 \\\hline
	\textbf{A1} [12.3]% \#8]
     & Within  & 8.8 & 7.9 & 7.8 & 	6.9& 	6.1& 	6.0& 	9.9& 	9.2& 	9.1 \\ 
	 & Between  &  15.2 & 14.2 & 14.0 &	16.9 &14.7 &	14.7 &	18.8 &	17.7 &	17.7 \\
    \multicolumn{2}{c}{}

\end{tabular}
~~~~
\begin{tabular}{ l  | l  l  l  l | l  l  l| l l l  }
	 \multicolumn{2}{c}{}&\multicolumn{3}{c}{Avg. Development}  &  \multicolumn{3}{c}{S1 -- German} & \multicolumn{3}{c}{S2 -- Wolof}  \\
	 \multicolumn{2}{c}{}&1 s&10 s& 2 min &  1 s & 10 s & 2 min  & 1 s & 10 s & 2 min \\ \hline
	\textbf{A2} [11.95]%: \#4]
     & Within & 8.8  & 7.8 & 7.7 & 	6.8& 	6.0& 	6.0& 	10.1& 	9.6 &	9.6 \\ 
	 & Between  & 14.5 & 13.4 & 13.3 &	16.0& 	14.0& 	13.9 &	17.9 &	16.9& 	16.6 \\	 
	\textbf{A3} [11:93]%: \#3]
     & Within  & 9.5 & 8.3 & 8.2 &	7.3 &	6.2 &	6.1 &	11.1 &	10.3 &	10.2\\ 
	 & Between & 14.5 & 13.3 & 13.2 &	15.5 &	13.5 &	13.4 &	17.6 &	16.0 &	16.0 \\
	\textbf{A4} [12.17]%: \#6]
     & Within  &  9.7 & 8.5 & 8.5 & 7.6 &	6.4 &	6.2 	&11.6 &	10.9 &	10.7\\ 
	 & Between   & 14.5 & 13.3 & 13.2 & 15.7&13.7&13.5&17.5&16.1&16.1 \\\hline
	 \textbf{Y1} [12.43]%: \#9]
     & Within  & 10.8 & 8.4 & 8.4 & 8.1 & 6.0 & 6.0 & 12.6 & 10.0 & 9.9\\ 
	 & Between  & 15.3 & 13 & 12.6 & 16.2 & 12.9 & 12.6 & 19.5 & 17.1 & 16.6\\
	 \textbf{Y2} [12.29]%: \#7]
     & Within  & 10.7 & 8.4 & 8.2 & 8.2 & 6.2 & 6.2 & 12.7 & 10.1 & 9.9\\ 
	 & Between  & 15.1 & 12.7 & 12.4 & 16.4 & 13.3 & 13.0 & 19.2 & 17.3 & 16.7\\\hline\hline
	\textbf{YS}  [12.32]
     & Within & 10.7 & 8.3 & 8.1 & 8.0 & 6.0 & 5.9 & 12.9 & 10.8 & 10.6\\ 
	 & Between & 14.7 & 12.3 &12.1 & 15.8 & 12.4 & 12.3 & 18.7 & 17.4 & 17.0\\\hline 
 	\textbf{S1} [8.54]
     & Within  & 8.7 & 8.3 & 7.3 & 6.3 & 5.8 & 5.0 & 9.0 & 8.7 & 7.2\\ 
	 & Between & 11.4 & 10.4 & 9 & 11.6 & 9.9 & 8.7 & 11.5 & 10.2 & 8.6 \\
	 	\textbf{S2} [6.99]
     & Within  & 7.1 & 6.7 & 6.3 & 5.2 & 4.9 & 4.5 & 6.9 & 7.0 & 6.3\\ 
	 & Between  & 9.0 & 8.6 & 7.8 & 9.3 & 8.6 & 7.8 & 8.3 & 7.9 & 7.2  \\

\end{tabular}
}
\end{table*}

%\begin{table*}[t]\label{tab:track2}
%\caption{Track 1 results. This table include only NED and Coverage, the full table is keep in our web-page. In the table C is the number of clusters and P the total number of pairs used in computing the scores}

%\footnotesize
%\begin{tabular}{ r |  r  r  r  r  r  r  r  r  r  r  r  r | r  r  r  r  r  r  r  r }

%     & \multicolumn{12}{c}{Development} & \multicolumn{8}{c}{Surprise} \\
%	 & \multicolumn{4}{c}{English} & \multicolumn{4}{c}{French} &  \multicolumn{4}{c}{Mandarin} & \multicolumn{4}{c}{L1} & \multicolumn{4}{c}{L2}  \\
%	 & C & P & NED & Cov & C & P & NED & Cov & C & P & NED & Cov & C & P & NED & Cov & C & P & NED & Cov \\ \hline
	 
%	\textbf{JHU} & 12886 & 15730 & 33.9 & 7.9 & 1803 & 1636 & 25.4 & 1.6 & 156 & 160 & 30.7 & 2.9 &
%   2973 & 3315 & 30.5 & 3.0 & 462 & 545 & 33.5 & 3.2 \\
%    \textbf{AG} & 8881 & 4222990 & 0.0 & 100.0 & 7215 & 3211559 & 0.0 & 100.0 & 1240 & 791707 & 0.0 & 100.0 
%    & _ & _ & _ & _ & _ & _ & _ & _ & \\
%   	\textbf{system1} & & & & & & & & & & & & & & & & & & & & & \\ \hline

%\end{tabular}
%\end{table*}

\section{Evaluation}
\label{sec:evaluation}

\subsection{Evaluation of subword modelling}

Unsupervised subword modeling is a representation learning problem: the system must find speech features that emphasize linguistically relevant properties (phoneme structure) and de-emphasize the linguistically irrelevant ones (speaker identity, emotion, channel, etc). We use a minimal pair ABX task \cite{schatz2013,schatz2014}, which does not require any training, and only requires a dissimilarity metric between speech tokens.  The ABX task is inspired by match-to-sample tasks used in human psychophysics and is a simple way to measure discriminability between two sound categories (where the sounds $A$ and $B$ belong to different categories $x$ and $y$, respectively, and the task is to decide whether the sound $X$ belongs to one or the other). We define the ABX-discriminability of category $x$ from category $y$ as the probability that $A$ and $X$ are further apart than $B$ and $X$ when $A$ and $X$ are from category $x$ and $B$ is from category $y$, according to a dissimilarity funcion $d$. We obtain a symmetric discriminability score by taking the average of the ABX discriminability of $x$ from $y$ and of $y$ from $x$. The dissimilarities provided in this challenge are based on dynamic time warping, the underlying frame-level dissimilarity being either the cosine distance or KL-divergence. For most systems (signal processing, embeddings) the cosine distance usually gives good results, and for others (posteriorgrams) the KL divergence is more appropriate. Participants are allowed to supply their own dissimilarity as long as it was not obtained through supervised training.

We focus on phone triplet minimal pairs: sequences of 3 phonemes that differ in the central sound (not necessarily real words, e.g., ``beg''--``bag'', ``api''--``ati'', etc). Our compound measure sums over all minimal pairs of this type found in the corpus in a structured manner, that depends on the task. For the \textit{within-speaker} task, all of the phone triplets belong to the same speaker (e.g. $A=\textrm{beg}_{T1}$, $B=\textrm{bag}_{T1}$, $X=\textrm{bag}'_{T1}$). The scores for a given minimal pair are first averaged across all of the speakers for which this minimal pair exists. The resulting scores are then averaged over all found contexts for a given pair of central phones (e.g. for the pair /a/-/e/, average the scores for the existing contexts such as /b\_g/, /r\_d/, /f\_s/, etc.). Finally the scores for every pair of central phones are averaged and subtracted from 1 to yield the reported within-talker ABX error rate. For the \textit{across-speaker} task, $A$ and $B$ belong to the same speaker, and $X$ to a different one. $A=\textrm{beg}_{T1}$, $B=\textrm{bag}_{T1}$, $X=\textrm{bag}_{T2}$. The scores for a given minimal pair are first averaged across all of the pairs of speakers for which this contrast can be made. As above, the resulting scores are averaged over all contexts over all pairs of central phones and converted to an error rate.

%As discussed above, track one systems are given training data, from which they must learn (in the case of the surprise languages) in a genuinely unsupervised way. The resulting representational transformation is computed for a test set consisting entirely of new talkers, divided into short files for which no talker labels are provided. The evaluation is done separately for three different lengths of audio files (two minutes, ten seconds, and one second) to assess the applicability of the system to new talkers with varying exposure lengths.

\subsection{Evaluation of spoken term discovery}

\newcommand{\set}[1]{\left\{#1\right\}}
\newcommand{\tup}[1]{\langle#1\rangle}
\newcommand{\cdisc}{C_{\textrm{disc}}}
\newcommand{\pdisc}{P_{\textrm{disc}}}
\newcommand{\fdisc}{F_{\textrm{disc}}}
\newcommand{\fall}{F_{\textrm{all}}}
\newcommand{\pall}{P_{\textrm{all}}}
\newcommand{\pgold}{P_{\textrm{all}}}
\newcommand{\psubs}{P_{\textrm{disc*}}}
\newcommand{\pclus}{P_{\textrm{clus}}}
\newcommand{\bdisc}{B_{\textrm{disc}}}
\newcommand{\bgold}{B_{\textrm{gold}}}
\newcommand{\pgoldclus}{P_{\textrm{goldclus}}}
\newcommand{\pgoldlex}{P_{\textrm{goldLex}}}
\newcommand{\fgoldlex}{F_{\textrm{goldLex}}}
\newcommand\SetB[3]{\ensuremath{\{\text{\ensuremath{#1 \mid} \parbox[t] {\widthof{\ensuremath{#3}}} {\ensuremath{#2}\}}}}}
\newcommand\SetA[2]{\ensuremath{\{\text{\ensuremath{#1 \mid} \parbox[t] {\widthof{\ensuremath{#2}}} {\ensuremath{#2}\}}}}}
\newcommand{\cover}[1]{\mathrm{cover}(#1)}
\newcommand{\flatten}[1]{\mathrm{flat}(#1)}
\newcommand{\nmatch}[1]{\mathrm{occ}(#1)} %% previously \#occ
\newcommand{\weight}[1]{\mathrm{weight}(#1)}
\newcommand{\types}[1]{\mathrm{types}(#1)}
\newcommand{\freq}[1]{\mathrm{freq}(#1)}
\newcommand{\ned}[1]{\mathrm{ned}(#1)}
\newcommand{\card}[1]{\left\vert{#1}\right\vert}

``Spoken term discovery'' is a broad label. We use it to refer to systems that do one or more of three tasks. Systems that do \emph{matching} find fragments that are matched pairwise as being instances of the same sequence of phonemes, attempting to find as many as possible. They can be evaluated based on how similar the matched fragments are, and how much of the corpus they cover. Systems that do \emph{lexicon discovery} group these fragments into sets (rather than just matching pairwise), with the goal of finding a high-quality lexicon of types. They can be evaluated based on how well the sets match on the sequence of phonemes, and how well the sets match the gold-standard lexicon of word types. Systems that do \emph{word segmentation} attempt to find fragments that are aligned with the gold-standard word transcription. By setting out three different types of criteria, the intention is to be open to various types of ``spoken term discovery'' systems, all of which in some sense ``find words.'' The result is that we do three (non-independent) types of evaluations in track two. All of these evaluations are done at the level of the phonemes. Using the aligned phoneme transcription, we convert any discovered fragment of speech into its transcribed string.  If the left or right edge of the fragment contains part of a phoneme, that phoneme is included in the transcription if it corresponds to more than more than 30ms or more than 50\% of its duration.

\subsubsection{Evaluation as matching system}

The evaluation of spoken term discovery systems as matching systems consists of two scores,  \textbf{NED} and  \textbf{coverage}. \textbf{NED,} is the average, over all matched pairs, of the Levenshtein distance between their phonemic transcriptions, divided by the max of their phonemic length.
%{\small 
%\begin{align}
%  d(\tup{i,j},\tup{k,l}) &= \frac{\textrm{Levenshtein}(T_{i,j}, T_{k,l})} {\max(j-i+1, k-l+1)}\label{eq:ned}
%\end{align}
%}
%The numerator is the standard Levenshtein distance, and $T_{i,j}$ is the transcription corresponding to the speech fragment designated by the pair of indices $[i,j]$ (the speech fragment between time points $i$ and $j$). 
The second score is the \textbf{coverage,} the fraction of the discoverable part of the corpus that is covered by all the discovered fragments. The discoverable part of the corpus is found by computing the union of all of the intervals corresponding to all of the pairs of ngrams (with n between 3 and 20). This is almost all of the corpus, except for unigram and bigram hapaxes.
\subsubsection{Evaluation as lexicon discovery system}

Six scores are used to evaluate the performance of a spoken term discovery system in terms of lexicon discovery. The first three are \textbf{grouping precision, recall} and \textbf{F-score.} These are defined in terms of $\pclus$, the set of all pairs of fragments that are groupes in the same cluster, and $\pgoldclus$, the set of all non-overlapping pairs of fragments which are both discovered by the system (not necessarily in the same cluster) and have exactly the same gold transcription. 

{\small
\begin{align}
 \mbox{Prec:} & \sum_{t\in\types{{\pclus}}} w(t,\pclus)\frac{\card{\nmatch{t, \pclus\cap\pgoldclus}}} {\card{\nmatch{t,\pclus}}} \label{eq:grouping:precision} \\
 \mbox{Rec:} & \sum_{t\in\types{{\pgoldclus}}} w(t,\pgoldclus)\frac{\card{\nmatch{t, \pclus\cap\pgoldclus}}} {\card{\nmatch{t, \pgoldclus}}}\label{eq:grouping:recall}
\end{align}
}

Where $t$ ranges over the types of fragments (defined by the transcription) in a cluster $C$, and $occ(t,C)$ is the number of occurrences of that type, $w$ the number of occurrences divided by the size of the cluster. In other words, Prec is a weighted measure of cluster purity and Rec, of the inverse of the cluster's fragmentation. The other three scores are \textbf{type precision, recall,} and \textbf{F-score}. Type precision is the probability that discovered types belong to the gold set of types (real words), whereas type recall is the probability that gold types are discovered. We restrict both sets to words between three and twenty segments long.

%\begin{align}
%  \mbox{type prec=} & & %\frac{\card{\types{\fdisc}\cap\types{\fgoldlex}}}{\card{\types{\fdisc}}}\label{eq:type:pre%cision} \\
%  \mbox{type rec=} & & %\frac{\card{\types{\fdisc}\cap\types{\fgoldlex}}}{\card{\types{\fgoldlex}}}\label{eq:type:%recall}
%\end{align}

\begin{table*}[ht]
%\begin{table}[H]
\caption{\label{tab:track2res} Track 2 results. Scores are averages over evaluations done on subparts of the corpus: the list of corpus gold utterances is split in groups of 1000 elements/utterances in order to compute the scores in parallel. Thus, the $F$-scores presented are not directly related to the precision and recall scores.}
\centering
\resizebox{1.3\columnwidth}{!}{
\begin{tabular}
%{ l | r r | S[table-format=3.1] | S[table-format=3.1] | S[table-format=3.1] S[table-format=3.1] S[table-format=3.1] | S[table-format=3.1] S[table-format=3.1] S[table-format=3.1] | S[table-format=3.1] S[table-format=3.1] S[table-format=3.1] | S[table-format=3.1] S[table-format=3.1] S[table-format=3.1] | S[table-format=3.1] S[table-format=3.1] S[table-format=3.1] }
{@{}c@{ }|@{ }c@{ }|@{ }c@{ }|@{ }c@{ }|@{ }c@{ }|@{ }c@{ }c@{ }c@{ }|@{ }c@{ }c@{ }c@{ }|@{ }c@{ }c@{ }c@{ }|@{ }c@{ }c@{ }c@{ }}

\multicolumn{5}{c}{} &  \multicolumn{3}{c}{Grouping}   & \multicolumn{3}{c}{Type}  & \multicolumn{3}{c}{Token}   & \multicolumn{3}{c}{Boundary}  \\
	 
& Words & Pairs & NED & Cov &  Pr & Re & F & Pr & Re & F & Pr & Re & F & Pr & Re & F  \\ \hline 

\multicolumn{17}{c}{ \textbf{Avg. Training (English, French, Mandarin)} } \\\hline
\textbf{Baseline}	&	4948	&	5842	&	30.0	&	4.1	&	48.7	&	86.6	&	52.3	&	5.5	&	0.3	&	0.6	&	4.4	&	0.2	&	0.2	&	34.1	&	1.5	&	2.9	\\
\textbf{Topline}	&	5779	&	2742085	&	0.0	&	100.0	&	99.9	&	100.0	&	100.0	&	41.3	&	45.9	&	43.4	&	47.8	&	58.4	&	52.2	&	77.9	&	94.4	&	84.9	\\\hline
\textbf{K}	&	24724	&	3775685	&	76.0	&	105.0	&	6.7	&	8.3	&	6.7	&	4.6	&	9.0	&	6.1	&	6.3	&	7.1	&	6.7	&	41.7	&	47.6	&	44.4	\\\hline
\textbf{G1}	&	51377	&	465997	&	71.5	&	60.5	&	3.9	&	67.8	&	7.4	&	3.9	&	6.0	&	4.4	&	3.1	&	2.7	&	2.9	&	25.3	&	31.8	&	27.8	\\
\textbf{G2}	&	51354	&	467369	&	71.4	&	60.4	&	4.0	&	67.6	&	7.5	&	3.9	&	6.0	&	4.4	&	3.1	&	2.7	&	2.8	&	25.2	&	31.8	&	27.7	\\

\multicolumn{17}{c}{ \textbf{S1 -- German} } \\\hline
\textbf{Baseline} & 2973 & 3315 & 30.5 & 3.0 & 54.8 & 94.6 & 64.9 & 5.5 & 0.3 & 0.6 & 4.0 & 0.1 & 0.2 & 28.2 & 1.2 & 2.3 \\
\textbf{Topline} & 7664 & 2588808 & 0.0 & 100.0& 100.0 & 100.0 & 100.0 & 28.4 & 29.5 & 29.0 & 42.3 & 62.7 & 50.5 & 70.0 & 98.3 & 81.8 \\\hline
\textbf{K} & 28675 & 4258731 & 66.4 & 100.0  & 11.8 & 5.9 & 7.9 & 5.7 & 11.2 & 7.5 & 10.3 & 14.3 & 12.0 & 42.6 & 56.5 & 48.6 \\\hline
\textbf{G1} & 60648 & 582009 & 59.9 & 71.8  & 5.7 & 63.8 & 10.4 & 3.2 & 6.8 & 4.3 & 2.4 & 3.1 & 2.7 & 20.6 & 37.2 & 26.6 \\
\textbf{G2} & 60489 & 588162 & 59.9 & 71.8  & 5.8 & 64.0 & 10.7 & 3.2 & 6.8 & 4.3 & 2.4 & 3.1 & 2.7 & 20.6 & 37.2 & 26.5 \\ \hline\hline

\multicolumn{17}{c}{ \textbf{S2 -- Wolof} } \\\hline
\textbf{Baseline} & 462 & 545 & 33.5 & 3.2 & 39.1 & 72.1 & 32.8 & 2.3 & 0.1 & 0.2 & 1.6 & 0.0 & 0.1 & 25.3 & 1.0 & 2.0 \\
\textbf{Topline} & 2472 & 602339 & 0.0 & 100.0  & 100.0 & 100.0 & 100.0 & 43.0 & 47.6 & 45.2 & 55.6 & 65.6 & 60.2 & 81.3 & 93.2 & 86.9 \\\hline
\textbf{K} & 3593 & 439321 & 72.2 & 100.0  & 4.7 & 5.4 & 5.0 & 4.6 & 10.0 & 6.3 & 4.9 & 5.2 & 5.0 & 42.4 & 44.3 & 43.3 \\\hline
\textbf{G1} & 5468 & 37191 & 56.8 & 47.8 & 7.6 & 43.3 & 12.8 & 5.9 & 8.2 & 6.9 & 6.1 & 4.2 & 4.9 & 29.6 & 29.6 & 29.6 \\
\textbf{G2} & 5460 & 37273 & 56.8 & 47.8  & 7.7 & 44.0 & 13.0 & 6.0 & 8.3 & 7.0 & 6.1 & 4.2 & 4.9 & 29.7 & 29.6 & 29.6 \\ 
\end{tabular}
}

\end{table*}
%\end{table}

\subsubsection{Evaluation as word segmentation system}

Six scores are used to evaluate the performance of a spoken term discovery system as a system for word segmentation. The first three are \textbf{token precision, recall,} and \textbf{F-score.} Token precision is the probability that discovered fragment tokens are in the gold set of word fragments, and token recall the probability that the gold fragments are discovered. %  These are defined with respect to $\fdisc$, the set of gold transcriptions corresponding to the discovered fragments (that is, the set of discovered fragment tokens), and $\fgoldlex$, the set of all word tokens according to the gold transcription. 
Again, tokens are restricted to be between three and twenty segments long.
%\begin{align}
%  \mbox{token precision=} & %\frac{\card{{\fdisc}\cap{\fgoldlex}}}{\card{{\fdisc}}}\label{eq:type:precision} \\
%  \mbox{token recall=} & %\frac{\card{{\fdisc}\cap{\fgoldlex}}}{\card{{\fgoldlex}}}\label{eq:type:recall}
%\end{align}
Finally, \textbf{boundary precision, recall,} and \textbf{F-score} are similarly defined with respect to the set of discovered and gold word boundaries.

\section{Description of submitted systems}
We received sixteen systems for track one, and three for track two. For two track one submissions, the authors declared that the systems were not intended as submissions, and are thus not included in the present overview, although the results can be found on the leaderboard on the Challenge website (\url{www.zerospeech.com/2017}).

\subsection{Track 1}

The baseline features are the MFCC features provided to participants; they are evaluated using the ABX score using the cosine similarity. The topline system is a supervised HMM-GMM phone recognition system, with a bigram language model, trained with a Kaldi recipe \cite{Povey_ASRU2011_2011}. The phone posteriors are evaluated using the KL divergence.

The submitted systems follow four strategies. The first strategy consists in performing a bottom-up frame-level clustering, following up on the success of \cite{chen15} with this simple strategy.
In system \textbf{[H]}, from \textbf{Heck et al.} \cite{heck}, PLP representations are used, as well as the output of three learned feature transformations, to be the input to the same frame-level clustering algorithm used in \cite{chen15} (the parallel Dirichlet process Gaussian mixture model sampler of \cite{pdpgmm}). The transformations (LDA, LDA+MLLT; LDA+MLLT+fMLLR) are aimed at making the representation talker invariant. The training requires labels, which are obtained from an initial clustering using the DPGMM algorithm. The submitted system is a 
combination of the raw posteriorgrams (not smoothed by any language model) from the four DPGMMs obtained by clustering frames in each of the three transformed representations, plus the raw PLP features.
System \textbf{[P1]}, from \textbf{Pellegrini et al.} \cite{pellegrini}, uses $k$-means clustering on ZCA-transformed (whitened using PCA and rotated back to the original axes) MFCC features. In system \textbf{[P2]}, this is followed by a step of centroid re-estimation: the centroids are re-estimated based only on the data points for which the label is the same as that of its left and right neighbors. The distance from point to all cluster centroids is the feature vector in both.

A second strategy constructs language-independent embeddings by training neural networks in an unsupervised way. The systems of \textbf{Chen et al.} \cite{chen} first cluster frames separately on each language, using the same DPGMM algorithm as \cite{heck} and \cite{chen15}. The cluster labels are then used as targets for a multitask neural network, where the loss function is the mean of the five cross-entropy losses. A linear bottleneck layer in the network is the final representation. System \textbf{[C1]} is trained on MFCCs; \textbf{[C2]} transforms the input features using unsupervised linear VTLN. The three systems in \textbf{Ansari et al.} \cite{ansari-ae} combine two sets of features, also trained on all five languages. The first set of features is always a high-dimensional hidden layer from an autoencoder trained on the MFCC frames. The second set of features is a hidden layer from a network trained to predict unsupervised frame labels. These labels are obtained by training a Gaussian mixture model on the speech frames. In system \textbf{[A1]}, the cluster labels are used as targets for the DNN trained with the MFCCs as input. System \textbf{[A2]} instead uses time-smoothed frame labels as decoded by a large Gaussian-mixture-HMM (one state for each of the initial cluster labels), in which states have a high self-transition probability. System \textbf{[A3]} does the same but trains the DNN with the autoencoder features as inputs. In systems \textbf{[A1]} and \textbf{[A2]}, the DNN features are combined with the autoencoder features in the dissimilarity function for the ABX task: the dissimilarity is a weighted average between the cosine DTW distance using the autoencoder features, and the cosine DTW distance using the other set of features (the weights are optimized on the development languages and not reported). An additional system is described in \cite{ansari-no-ae}, system \textbf{[A4]}, which consists of the DNN features from system \textbf{[A2]}.

The third strategy is to use spoken term discovery to improve the acoustic features, following \cite{thiolliere,renshaw}. The system of \textbf{Yuan et al.} \cite{yuan} generates bottle-neck features in the same way as \cite{chen}, but then applies the STD system of \cite{aren} to discover pairs of matched acoustic motifs, which are then used to further train the features. Specifically, following \cite{renshaw}, a stacked autoencoder is trained on DTW-aligned frame pairs from instances of motifs paired by the STD system. System \textbf{[Y1]} uses pairs obtained by applying the STD system (only) to the English training data. System \textbf{[Y2]} (not reported in \cite{chen}) obtained pairs from all corpora of all five languages. 
System \textbf{[YS]} is a supervised comparison, using transcribed pairs from the Switchboard corpus rather than STD.

%\subsubsection{Räsänen et al's system}
%Syllable-based pattern discovery system. 
%
%Track1 first extracts MFCC features (+delta +deltadelta) and applies mean and variance normalization to them. In order to acquire
%low-dimensional frame-level features for track2, the system then applies k-means clustering to the features and uses cluster indices as labels in linear discriminant analysis (LDA) for the MFCCs. These new features are then subjected to new clustering (with initialization from the previous iteration) and new LDA, and the process is repeated until the process converges to a point where the LDA can no longer improve separability of the clusters. On every iteration, feature dimensions with small eigenvalues are dropped out to compress the data.  
%
%See Heck et al. (2016), "Unsupervised Linear Discriminant Analysis for Supporting DPGMM Clustering in the Zero Resource Scenario", for the original idea where they used DPGMM + LDA on stacked MFCCs to improve quality of GMM posteriors.  
%
%The present approach is somewhat worse than the GMM posteriors from Heck et al. (2016), but provides a compact low-dimensional MFCCs with better distributional separability than standard MFCCs with CMVN. Importantly, the k-means+LDA -process tries to (implicitly) optimize feature distributions to become hyperspherical in Euclidean space, making the features potentially useful and computationally tractable for track2.

The strategy of \textbf{Shibata et al.} \cite{shibata} is to use supervised training on out-of-domain languages. In \textbf{[S1]}, the features are bottleneck features from a neural network acoustic model trained supervised on Japanese as part of an HMM. In \textbf{[S2]}, an end-to-end convolutional network plus bidirectional LSTM is trained using corpora from ten languages, and features from a hidden layer are combined with the bottleneck Japanese features. This second scheme can only be considered out of domain in a weak sense: new corpora are used, but the ten additional languages include English, Mandarin, and German.

\subsection{Track 2}

The baseline of track two was computed using \cite{aren}, which does pair-matching using locally sensitive hashing applied to PLP features and then groups pairs using graph clustering. The parameters of the STD stayed the same across all languages, except that the DTW threshhold was increased for Mandarin (to 0.90, from 0.88) in order to obtain a NED value similar to that of other languages. The topline was an exhaustive-parsing word segmentation model based on the textual transcriptions (a unigram grammar trained  in the adaptor grammar framework: \cite{adaptorgrammar}).

%\subsubsection{Räsänen et al's system}
%Syllable-based pattern discovery system. 
%
%Track2 uses envelope-based syllabification of the data, followed by detection of N (here: N = 3) best matching pairs for each syllable using a fixed-dimensional feature representation (downsampled track1 features) for each syllable and euclidean distance as the metric. Alignments between these best matching candidates are then analyzed more closely with DTW with a quality criterion that has a controllable trade-off between pattern length and average distance. In the end, pairs with pairwise average distance smaller than a pre-defined threshold and length larger than minimum (here: 50 ms) are stored as outputs for evaluation.
%
%The total execution time for the Mandarin corpus is roughly two hours with the given hyperparameters with the machine described above. For other corpora, that is much more as I ran out of time to optimize anything (or, TBH, to even develop track2 properly).

System \textbf{[K]}, from \textbf{Kamper et al.} \cite{kamper} uses $k$-means to discover recurring acoustic patterns, jointly optimized with an exhaustive segmentation. The variable-length acoustic chunks of speech yielded by a segmentation are reduced to fixed dimension by down-sampling in the time dimension for clustering by $k$-means. The systems of \textbf{Garc\'ia-Granada et al.} \cite{garcia} use a supervised ASR system for Hungarian to decode the speech. Chunks of (transcribed) speech match if they have the same transcription. These matches are then filtered: a representation of the speech is obtained by training an autoencoder, and only pairs with DTW sufficiently low in this representational space (below a threshhold) are retained (system \textbf{[G1]}). In system \textbf{[G2]}, pairs are tolerated that have a one-phone difference in their transcriptions.

\section{Results}
\label{sec:results}

%For each of the submitted subword models, plus the two baselines described above (acoustic baseline and supervised topline), the features are computed for the test set, for each corpus. We perform the ABX evaluations described above (results on development languages are those computed and reported by the authors; the surprise language evaluations are computed by the conference organizers). 
The results of track one are summarized in Table \ref{tab:track1res}. 
The clear winner among the unsupervised models is \textbf{[H]}. While the paper focuses on the role of learned feature transformations---and demonstrates that they improve the evaluation---these are unlikely to be solely responsible for the very good performance of the system. The baseline system described in the paper, with no transformations applied, consisting of posteriorgrams from a frame-wise DPGMM clustering, still performs extremely well. Although it was not evaluated on the surprise languages, its average ABX score on the development languages (9.7) is still better that of the second-ranked system \textbf{[C2]} (10.7), from \textbf{Chen et al.} The systems presented by \textbf{Heck et al.} all have in common monolingual training, direct evaluation of GMM posteriorgrams, and the fact that they use PLP features as input, rather than MFCCs.

The systems of \textbf{Chen et al.} (as well as \textbf{Yuan et al.}) use an initial stage of monolingual training, followed by multilingual training. Among the other systems in \cite{chen} (not submitted for final evaluation) are some with fully monolingual training, which serve to partly evaluate the effect of multilingual training: monolingual training fares better, within language. However, as the numerical scores are not presented, it is not possible to determine whether the use of monolingual training could account for the good scores of \textbf{Heck et al.} What is clear is that multilingual training still works reasonably well---all the remaining systems, except for the lower performing system \textbf{[P]}, use multilingual training, and show substantial improvements over the baseline. The setting for these systems corresponds to the situation of an infant raised in a multilingual environment.

The STD strategy for improving subword features fares relatively poorly; a limitation of the \textbf{Yuan et al.} systems is that they use an architecture (autoencoders) that was previously reported to be suboptimal for improving acoustic features \cite{renshaw}. A Siamese network \cite{thiolliere} might improve the result.

The results of track two are in Table \ref{tab:track2res}. 
%For each of the submitted spoken term discovery models, plus the two baselines described above (the PLP-LSH based system and the exhaustive text-based system), the list of classified pairs is computed for the entire subsampled evaluation set, for each corpus. We calculate the scores described above (results on development languages are computed and reported by the authors; the surprise language evaluations are computed by the conference organizers). 
As in the previous challenge, the baseline provided by the system of \cite{aren} has a high NED (very good matching), but has a low coverage (few matches). Given that system \textbf{[K]} demands an exhaustive parse, it is guaranteed to have full coverage, which risks poorer matches. It is thus notable that system \textbf{[K]} has a NED on the development languages comparable to the partly supervised systems \textbf{[G1]} and \textbf{[G2]}. On the other hand, grouping by supervised transcription clearly works better than $k$-means clustering; while the groups have low precision (many erroneously grouped items) for both kinds of systems, \textbf{[G1]}/ \textbf{[G2]} have much higher grouping recall (more of the true group members among the items grouped together) than system \textbf{[K]}. The quality of segmentation, in terms of tokens and of boundaries, is, on the other hand, much better in system \textbf{[K]}, presumably as a result of increased coverage.

\section{Conclusion}
\label{sec:conclusion}

The results of the 2017 Zero Resource Speech Challenge build very clearly on the previous challenge. The strategies most successful in the previous challenge for track one, bottom-up clustering and training based on pairs of matching words discovered using spoken term discovery---are built upon, and a new strategy, multilingual training, is introduced. This strategy, although not completely foreseen by the organizers, would be similar to the situation of a multilingual infant. This new approach is used in conjunction with additional tools, including further training using neural networks and the application of time series models.  Interestingly, simple bottom-up clustering still seems to work best in the unsupervised case.  The fact that supervised labels trained fully out of domain seem to yield good results in both track one and track two (on Japanese, in the case of \textbf{[S1]}, and on Hungarian in the case of \textbf{[G1]} and \textbf{[G2]}) indicates that the approach of having systems with strong prior knowledge about human speech is still a benchmark to be beaten in the unsupervised case. (The excellent results of \textbf{[S2]} may yet be partly attributable to the use of supervised training in German, which was one of the surprise languages.) Further work is needed to systematically evaluate the effect of each of these different manipulations, however, and to directly compare these new approaches with those from the previous challenge. 

Potential future lines of inquiry building on the new track two systems might include exploring using these systems to feed subword modelling, especially in combination with some of the new subword modelling strategies used here, in light of the successful use of STD to improve subword models in the previous challenge, and the reasonable improvement over baseline shown in this challenge. In the other direction, the major unsupervised improvements to the speech features demonstrated in track one suggest strong room for improvement in unsupervised spoken term discovery systems like \textbf{[K]}.

The 2017 ZeroSpeech Challenge remains open for submissions exploring these, or any other new strategies for discovering phoneme- and word-like units without access to linguistic resources.

\bibliographystyle{IEEEbib}
\clearpage
\bibliography{refs}

\end{document}